%% file: placing_draft.tex
\documentclass[conference]{IEEEtran}
\usepackage{times}

\usepackage[numbers]{natbib}
\usepackage{array}
\usepackage{relsize}
\usepackage{multirow}
\usepackage{subfig}
\usepackage{amsmath}
\usepackage{amsfonts, bm}
\usepackage[bookmarks=true]{hyperref}
\usepackage{graphics} 
\usepackage{graphicx}
\usepackage{booktabs}

\usepackage{color}

\newlength\savedwidth
\newcommand\whline[1]{\noalign{\global\savedwidth\arrayrulewidth
                               \global\arrayrulewidth #1} %
                      \hline
                      \noalign{\global\arrayrulewidth\savedwidth}}

\renewcommand{\Re}{\mathbb{R}}

\begin{document}

\title{Learning to Place New Objects}

\author{

Yun Jiang, Changxi Zheng, Marcus Lim and Ashutosh Saxena\\
Computer Science Department, Cornell University.\\
\{yunjiang,cxzheng\}@cs.cornell.edu, mkl65@cornell.edu, asaxena@cs.cornell.edu 
}



%

\maketitle

\input{abstract}

\IEEEpeerreviewmaketitle

\input{introduction}

\input{related_work}

\input{system_overview}

\input{feature_extraction}
\input{algorithm}

\input{experiment}
\input{discussion}
\input{conclusion}



{\small
\bibliographystyle{IEEEtran}
\bibliography{references}
}

\end{document}

%% file: abstract.tex
\begin{abstract}
The ability to place objects in an environment is an important skill for a personal robot.
An object should not only be placed stably, but should also be placed in its preferred location/orientation.
For instance, it is preferred that a plate be inserted vertically into the slot of a dish-rack
as compared to being placed horizontally in it. 
Unstructured environments such as homes have a large variety of object types as well as  
of placing areas. Therefore our  algorithms should  be able to handle placing 
 \textit{new} object types and \textit{new} placing areas.  
These reasons make placing a challenging manipulation task.

In this work, we propose using supervised learning approach for finding good
placements given the point clouds of the object and the placing area. It 
learns to combine the features that capture support, stability and preferred placements
using a shared sparsity structure in the parameters.
Even when neither the object nor the placing area is seen previously in the training set,
our algorithm predicts good placements.
In extensive experiments, 
our method enables the robot to stably place several new objects in several new placing areas
with a 98\% success-rate, and it placed the objects
in their preferred placements in 92\% of the cases.

\end{abstract}

%% file: introduction.tex
\section{Introduction}

In several manipulation tasks of interest, such as arranging a disorganized kitchen,
loading a dishwasher or laying a dinner table, a robot needs to pick up
and place objects.
While grasping has  attracted great attention in previous works, placing remains
under-explored.  To  place objects successfully, a robot needs to figure
out where and in what orientation to place them---even in cases when the objects and
the placing areas may have not been seen  before.

\begin{figure}[t]
\begin{centering}
        \includegraphics[height=1.70in]{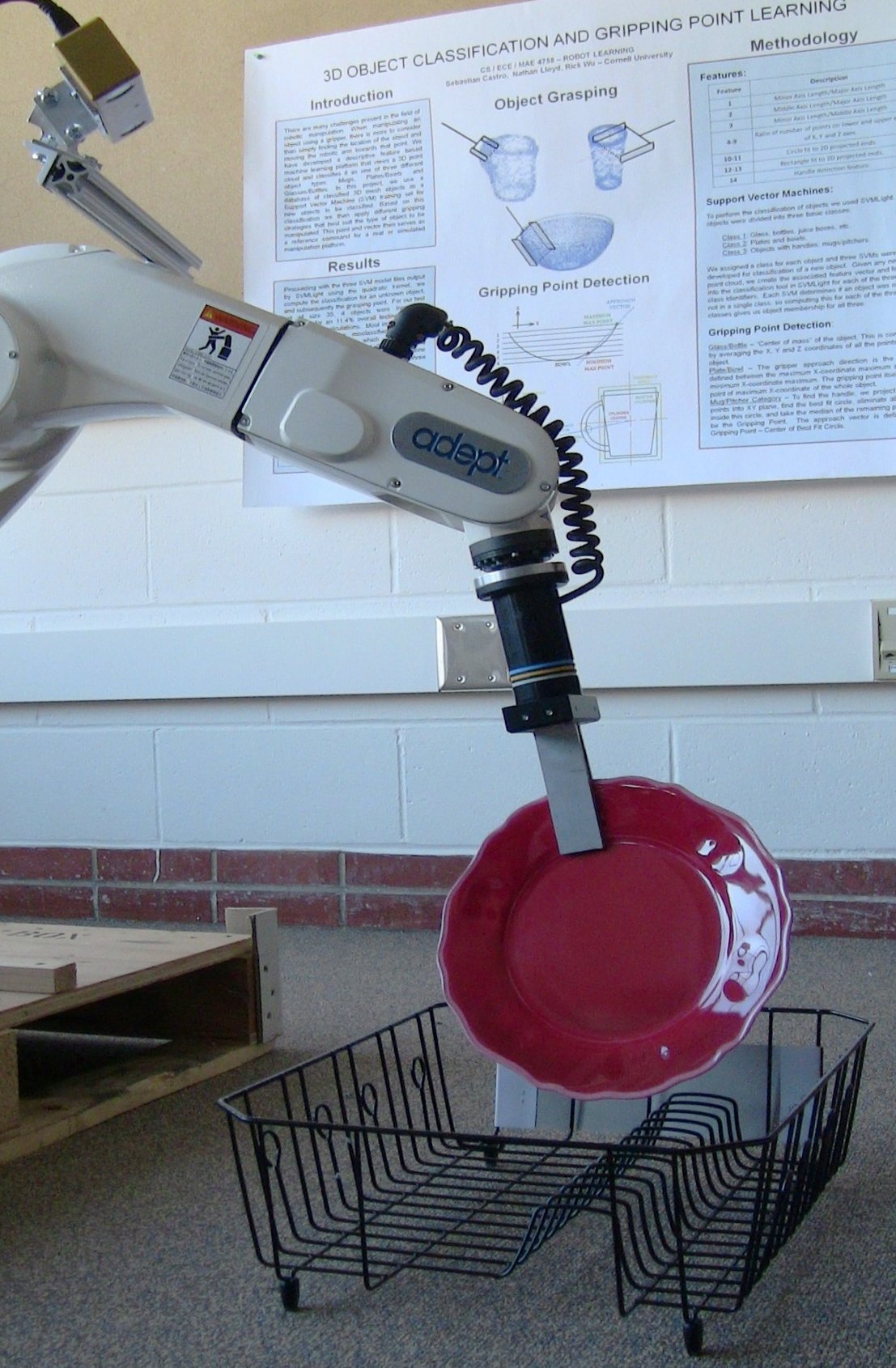} 
		\includegraphics[height=1.70in]{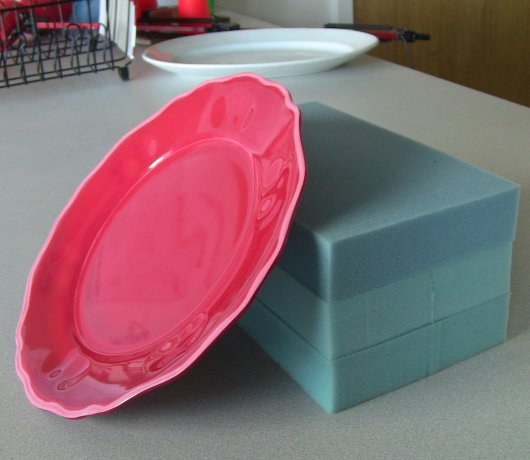} 
\caption{{\small How to place an object depends on the shape of the object and
the placing environment.  For example, a plate could be placed vertically in
the dish rack (left), or placed slanted against a support (right).
Furthermore, objects can also have a `preferred' placing configuration. E.g.,
in the dish rack above, the preferred placement of a plate  is vertically into the
rack's slot and not horizontally in the rack.}
} 
\label{fig:placing}
\end{centering}
\end{figure}

Given a designated placing area (e.g., a dish-rack),
this work focuses on finding good placements (which includes the location and
the orientation) for an object.
An object can be placed stably in several different ways---for example, a plate
could be placed horizontally on a table or placed vertically in the slots of a
dish-rack, or even be side-supported when other objects are present (see
Fig.~\ref{fig:placing}). A martini glass should be placed upright on a table but upside
down on a stemware holder. 
In addition to stability, some objects also have `preferred' placing configuration that can be learned
from prior experience.  For example, long thin objects (e.g., pens, pencils,
spoons) are placed horizontally on a table, but vertically in a pen- or  
cutlery-holder. Plates and other `flat' objects are placed horizontally on a table, but
plates are vertically inserted into the `slots' of a dish-rack.
Thus there are certain common features depending on the shape of objects and 
placing areas that indicate their
preferred placements.  These reasons make the space of potential placing
configurations of common objects in indoor environments very large.
The situation is further exacerbated when the robot has not seen the
object or the placing area before. 


In this work, we compute a number of features
that indicate stability and supports, and rely on supervised learning techniques
to learn a functional mapping from these features to good placements. 
We learn the parameters of our model by maximizing the margin between the 
positive and the negative class (similar to SVMs~\cite{SVM}).  However we note that while
some features remain consistent across different objects and placing areas,
there are also  features that are specific to particular objects and
placing areas. We therefore impose a shared sparsity structure in the parameters
while learning.


\begin{figure*}[t!]
\centering
\includegraphics[height=0.095\textheight]{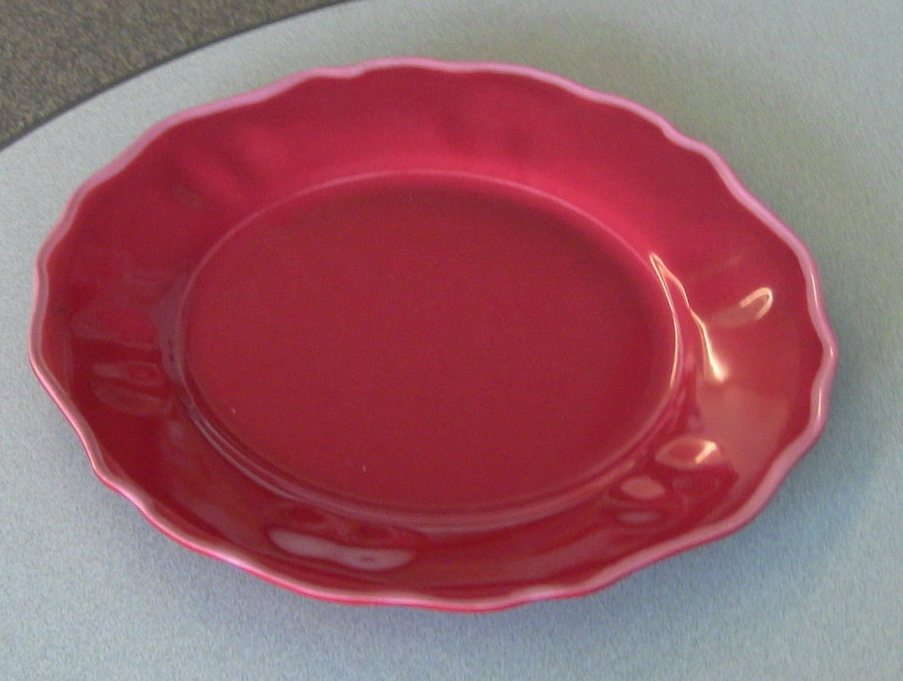}
\includegraphics[height=0.095\textheight]{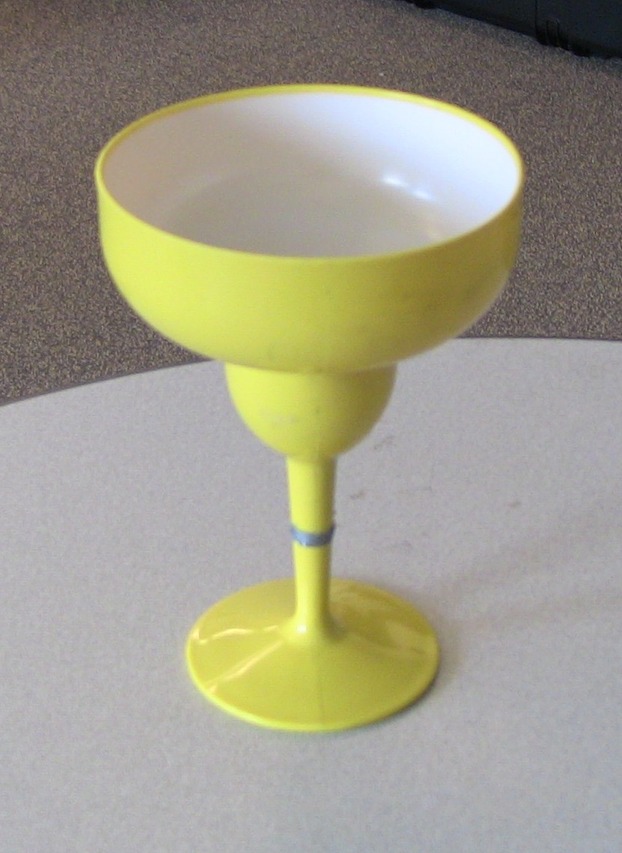}
\includegraphics[height=0.095\textheight]{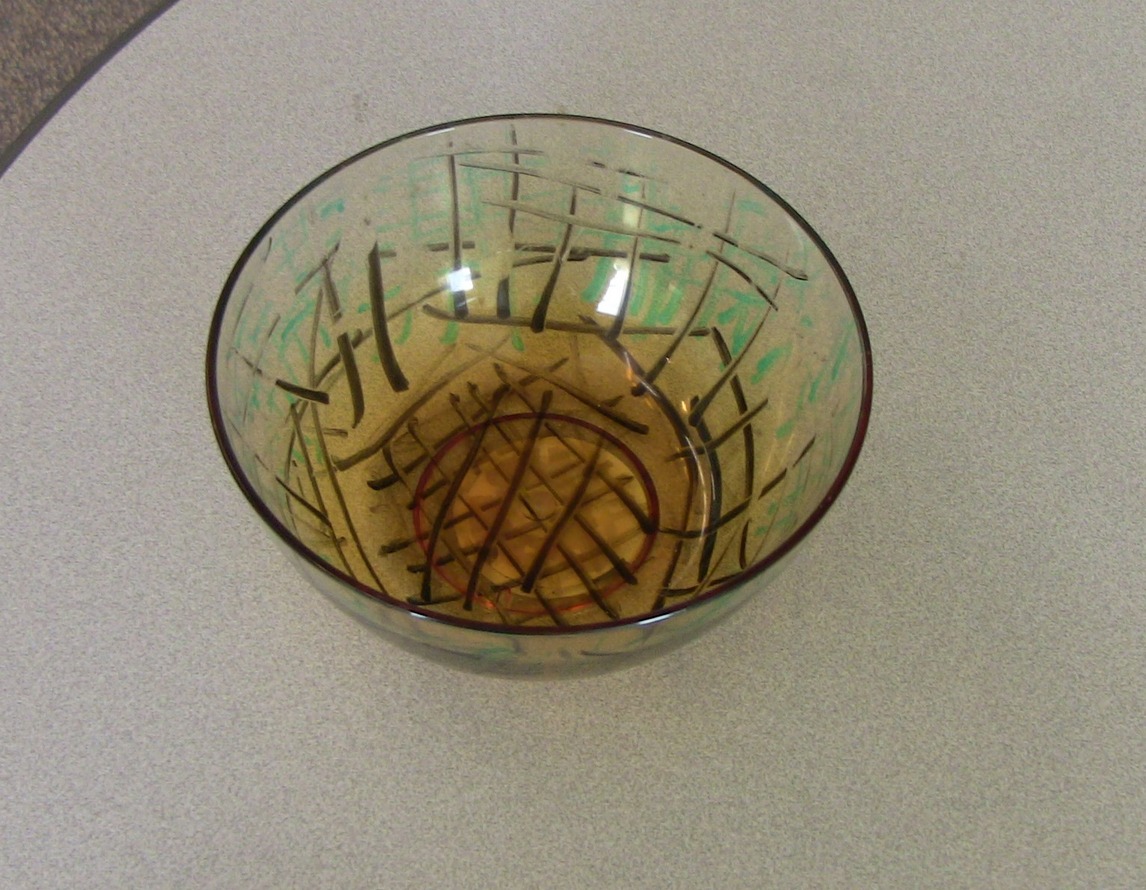}
\includegraphics[height=0.095\textheight]{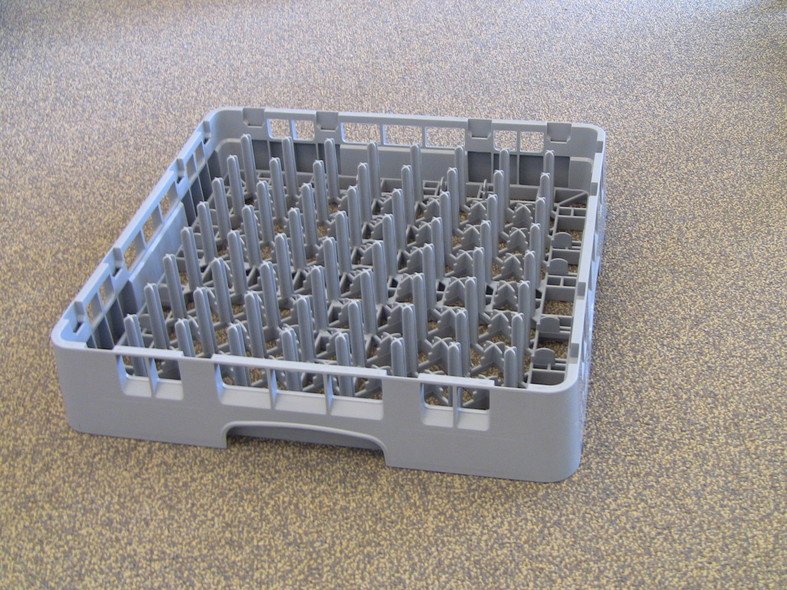}
\includegraphics[height=0.095\textheight]{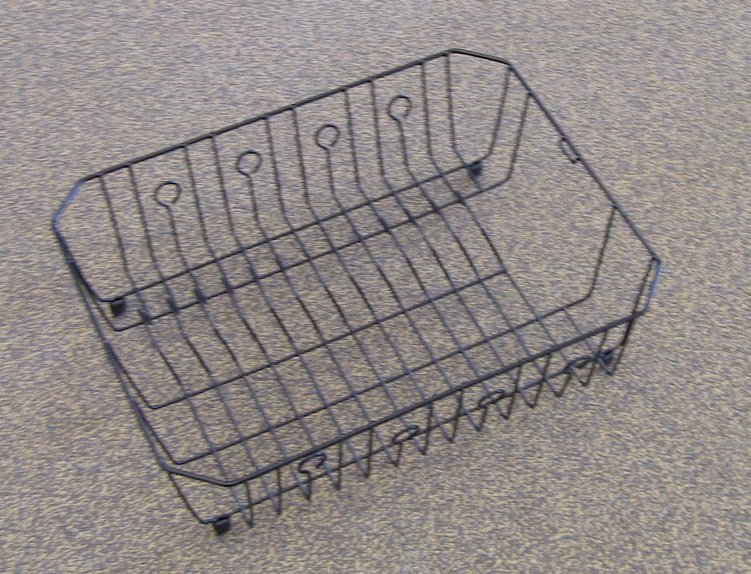}
\includegraphics[height=0.095\textheight]{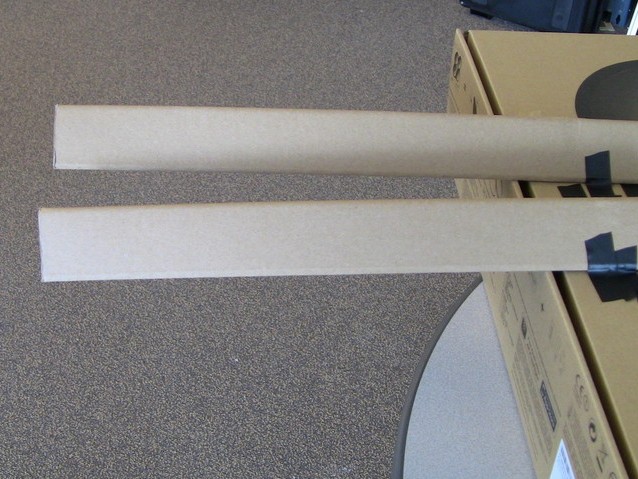}
\caption{{\small Examples of a few objects (plate, martini glass, bowl) and placing areas (rack-1, rack-3, stemware holder). For the full list, see Table~\ref{tbl:offline_results}.}}
\label{fig:introduction}
\end{figure*}

For training our model, we obtain ground-truth labeled data using 
rigid-body simulation. During robotic experiments, 
we first use a pre-constructed database of 3D models of objects to recover 
the point cloud, and then 
evaluate the potential placements using the learned model 
for obtaining a ranking. 
We next validate the top few placements 
in a rigid-body simulator (which is computationally expensive), and perform
 the placements using our robotic arm.

We test our algorithm extensively in the tasks of placing several objects
in different placing environments. (See Fig.~\ref{fig:introduction}
for some examples.) The scenarios range from simple placing of objects on flat
surfaces to narrowly inserting plates into dish-rack slots to hanging martini
glasses upside down on a holding bar.
We perform our experiments with a robotic arm that has no tactile
feedback (which makes good predictions of placements important). 
Given an object's grasping point and the placing area,
our method enables our robot to stably place previously unseen objects in
new placing areas with 98\% success-rate.  Furthermore, the objects 
were  placed in their \textit{preferred} configurations in 98\% of the cases when
the object and placing areas were seen before, and in 92\% of the cases when
the objects and the placing areas were new.

\medskip

The contributions of this work are as follows:
\begin{itemize}
\item While some prior work studies finding `flat' surfaces 
\cite{kemp-placing}, we believe that this is the first work that considers
placing new objects in complex 
placing areas. 
\item Our learning algorithm captures features that indicate stability
of the placement as well as the `preferred' placing configuration for an object. 


\end{itemize}

\bigskip

%% file: related_work.tex
\section{Related Work}\label{sec:related-work}


There is little previous work in placing objects, and it is restricted to
placing objects upright on `flat' surfaces.  Edsinger and Kemp
\cite{edsinger2007manipulation} considered placing objects on a flat shelf. The robot
first detected the edge of the shelf, and then explored the area with its arm
to confirm the location of the shelf. It used passive compliance and force
control to gently and stably place the object on the shelf. 
This work  indicates that even for flat surfaces,
unless the robot knows the placing strategy very precisely, it takes good tactile sensing and 
adjustment to implement placing without knocking the object down. Schuster et al.~\cite{kemp-placing}
recently developed a learning algorithm to
detect clutter-free `flat' areas where an object can be placed. 
While these works assumes that the given object is already in its upright orientation,
some other related works consider how to find the upright or the current orientation of the objects, e.g.
Fu et al.~\cite{siggraph_upright} 
proposed several geometric features to learn the upright orientation from an object's 3D
model and Saxena et al.~\cite{saxena_orientation} predicted the orientation of an object
given its image.
Our work is different and complementary to these studies: we generalize placing
environment from flat surfaces to more complex ones, and desired configurations are
extended from upright to all other possible orientations that can make the best use of the
placing area. 

Planning and rule-based approaches have been used in the past to move objects around.
For example, Lozano-Perez et al.~\cite{lozano2002task} considered picking up and placing 
objects by decoupling the planning problem into parts, and tested on grasping
objects and placing them on a table.  Sugie et al.~\cite{sugie2002placing} used
rule-based planning in order to push objects on a table surface.
However, these approaches assume known full 3D model of the objects,
consider only flat surfaces, and do not model preferred placements. 

In a related manipulation task, grasping, 
learning algorithms have been shown to be promising.
Saxena et al. \cite{SaxenaIJRR, SaxenaAAAI,saxena_thesis}
used supervised learning with images and partial point clouds as inputs to infer good grasps. 
Later, Le et al.~\cite{le2010learning} and Jiang et al.~\cite{jiangICRA} proposed 
new learning algorithms for grasping. Rao et al. \cite{rao-grasping} used point cloud to
facilitate segmentation as well as learning grasps.  Li et al.~\cite{feccm_grasping} combined
object detection and grasping for higher accuracy in grasping point detection.
In some grasping works that
assume known full 3D models (such as GraspIt \cite{GraspIt}), different 3D
locations/orientations of the hand are evaluated for their grasp scores.  
Berenson et al.~\cite{berenson2009grasp} consider grasping planning in complex
scenes. Goldfeder~\cite{goldfeder2009data} recently discussed a data-driven grasping approach.
In this paper, we consider learning placements of 
objects in cluttered placing areas---which is a different problem because
our learning algorithm has to consider the object as well as the environment
 to evaluate good placements.  To the best of our knowledge, we have not
seen any work about learning to place in complex situations.

%% file: system_overview.tex
\begin{figure*}[t!]
\centering
\includegraphics[height=0.083\textheight]{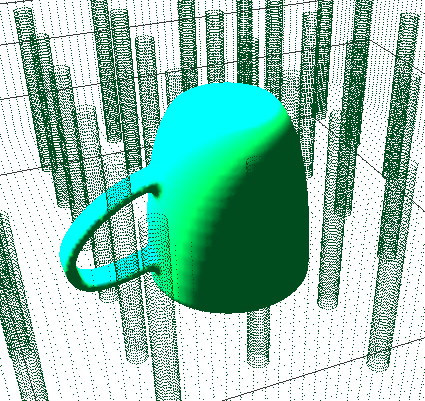}
\includegraphics[height=0.083\textheight]{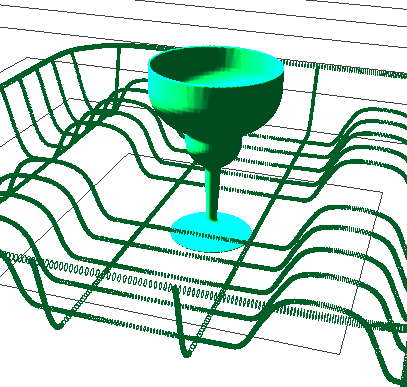}
\includegraphics[height=0.083\textheight]{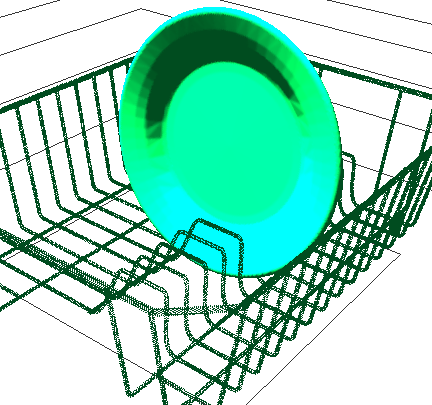}
\includegraphics[height=0.083\textheight]{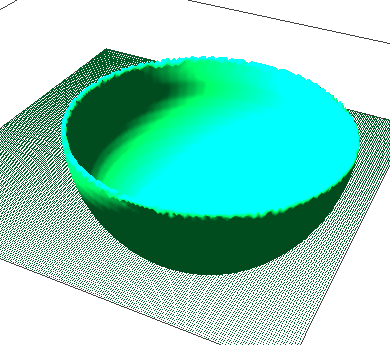}
\includegraphics[height=0.083\textheight]{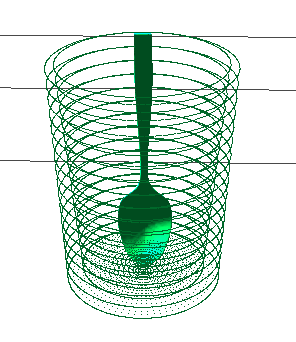}
\includegraphics[height=0.083\textheight]{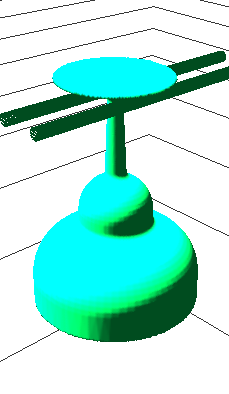}
\includegraphics[height=0.083\textheight]{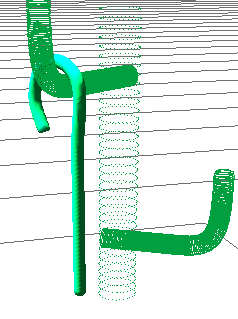}
\includegraphics[height=0.083\textheight]{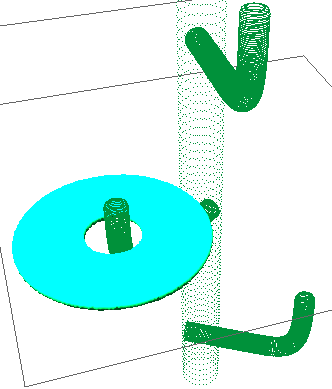}
\includegraphics[height=0.083\textheight]{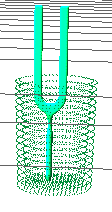}
\caption{{\small Some snapshots from our rigid-body simulator showing different objects
placed in different placing areas. (Placing areas from left: rack-1, rack-2, rack-3, flat surface, pen holder,
stemware holder, hook, hook and pen holder. Objects from left: mug, martini glass, plate, bowl, spoon, martini glass, candy cane, disc and tuning fork.)}}\label{fig:4_scenes}
\end{figure*}

\section{System Overview}\label{sec:system}
\begin{figure}[t!]
\centering
\includegraphics[width=1.0\columnwidth]{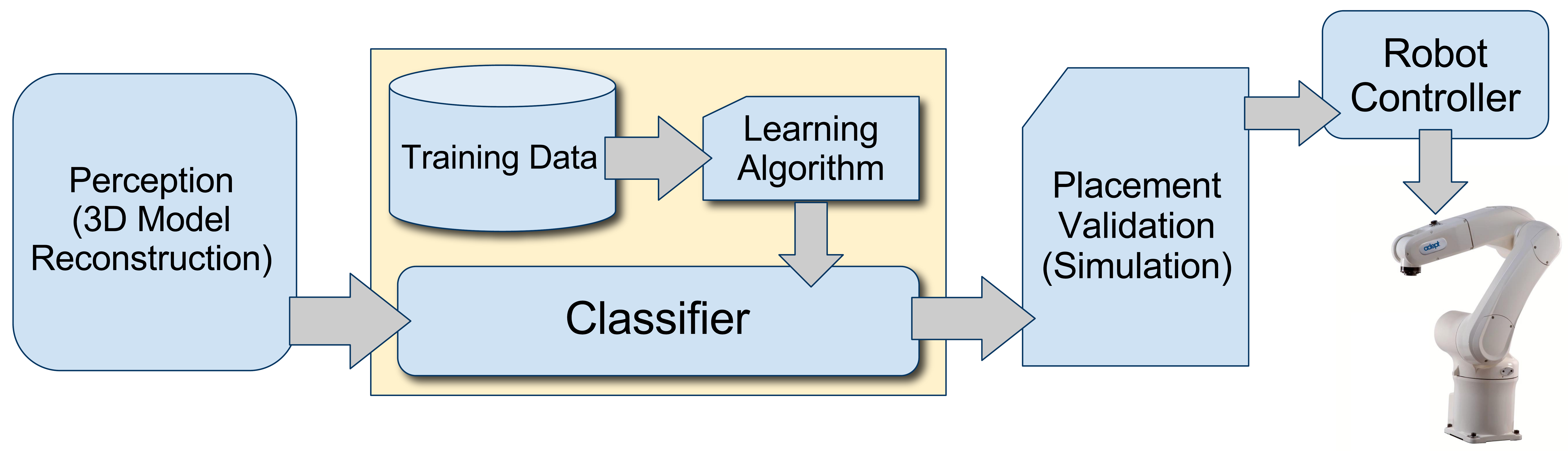}
\caption{ {\bf System Overview:} The core part is the placement classifier (yellow box) which is trained
using supervised learning to identify a few candidates of placement configurations based on
the perceived geometries of environments and objects. Those candidates are validated by 
a rigid body simulation to determine the best feasible placement which is then fed to the robot controller.
\label{fig:overview}}
\end{figure}

As outlined in Fig.~\ref{fig:overview}, the core part of our system 
is the placement classifier with supervised learning (yellow box in Fig.~\ref{fig:overview}), 
which we will describe in details in section~\ref{sec:algorithm}. In this section, we briefly 
describe the other parts.




\subsection{Perception}\label{sec:perception}
We use a stereo camera to perceive objects and placing areas, however,
it can capture the point cloud  only partially due to the 
shiny/textureless surfaces or occlusions. To recover the entire geometry, a
database of parameterized objects with a variety of shapes is created beforehand.
A scanned partial point-cloud is registered against the objects in the database using
the Iterative Closest Point (ICP) algorithm \cite{ICP1,ICP2}. The best matching object
from the database is used to represent the completed geometry of the scanned object.
While this 3D recovery scheme is not compulsory, with our particular stereo camera we have
found that it significantly improves the performance. In this work,
our goal is to study placement, therefore we simplify perception by assuming 
a good initial guess of the object's initial location.

\subsection{Simulation}\label{sec:simulation}

In our pipeline, rigid-body simulation
is used for two purposes: (a) to
generate training data for our supervised learning algorithm, and (b) to verify
the predicted placements suggested by the classifier (See Fig.~\ref{fig:4_scenes}
for some simulated placing tasks).

A placement defines the location $\bm{T}_0$ and orientation $\bm{R}_0$ of the object in the environment.
Its motion is computed by the rigid-body simulation at discrete time steps. At each time step,
we compute the kinetic energy change of the object $\Delta E = E_n-E_{n-1}$. The simulation
runs until the kinetic energy is almost constant ($\Delta E<\delta$), in which a stable state of
the object can be assumed. Let $\bm{T}_s$ and $\bm{R}_s$ denote the stable state. We label the
given placement as a valid one if the final state is close enough to the initial state, i.e.
    $\|\bm{T}_s - \bm{T}_0\|_2^2 + \|\bm{R}_s - \bm{R}_0\|^2_2 < \delta_s$.

This simulation is computationally expensive. 
Thanks to our classifier, we only need to perform the rigid-body simulation on a few suggested placements, 
thus making it a much more efficient process as compared to checking all random placements. 

Since the simulation itself has no knowledge of placing preferences, when
creating the ground-truth training data, we manually labeled all the
stable (as verified by the simulation) but un-preferred placements as negative examples.

\subsection{Realization}

To realize a placement decision, our robotic arm grasps the
object, moves it to the placing location with desired orientation, and releases it.
Our main goal in this paper is to learn how to place objects well, 
therefore in our experiments the kinematically feasible grasping points are
pre-calculated. The location and geometry of the object and the environment is computed from
real point cloud, using the registration algorithm in section \ref{sec:perception}.
The robot then uses inverse kinematics to figure out the arm configuration and plans a path
to take the object to the predicted placing configuration (including
location and orientation), 
 and then releases it for placing.



%% file: feature_extraction.tex
\begin{figure*}[t] \centering
    \subfloat[supporting contacts] {\includegraphics[width=0.45\columnwidth]{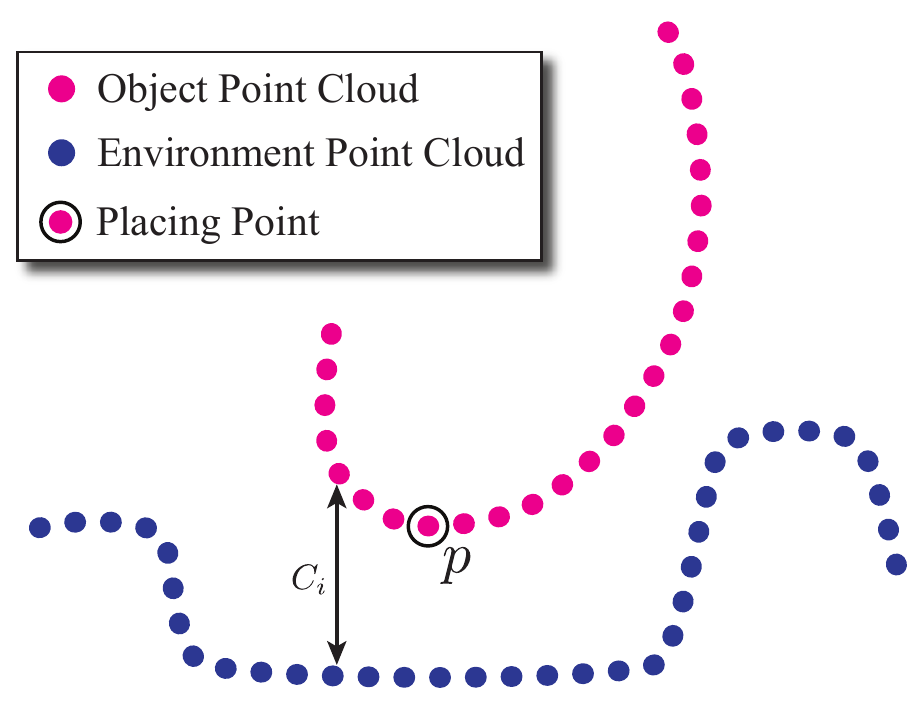}\label{fig:features_a}}
    \qquad
    \subfloat[caging (side view)] {\includegraphics[width=0.45\columnwidth]{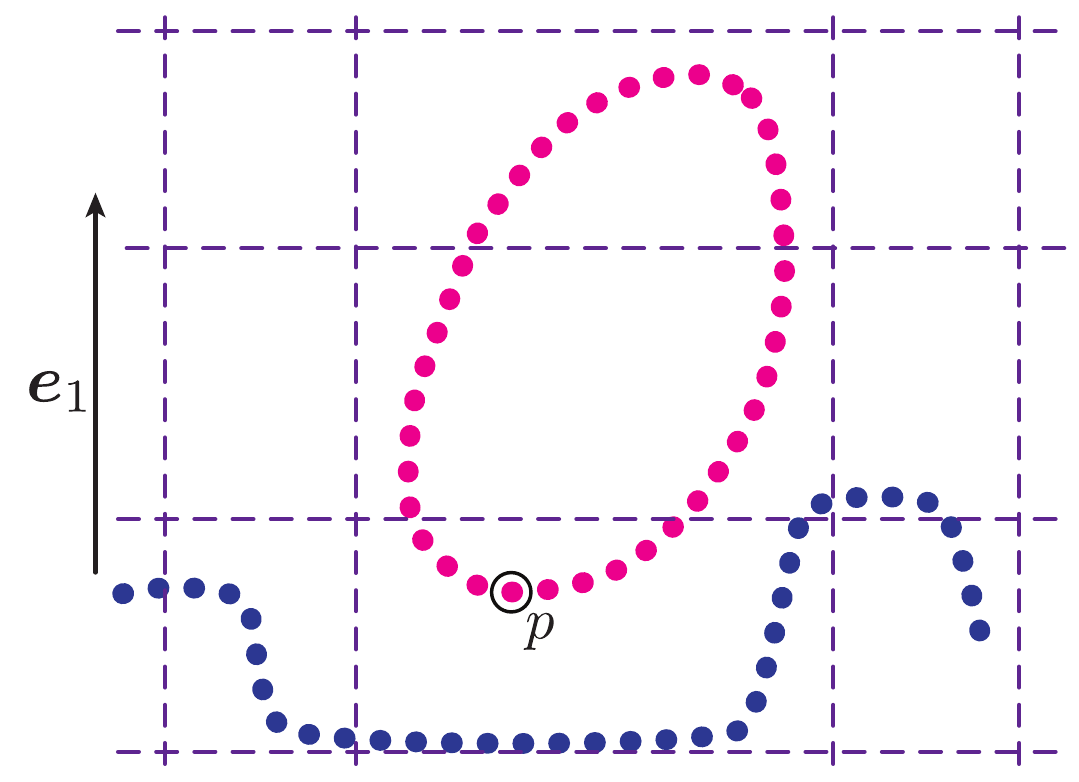}\label{fig:features_b}}
    \qquad
    \subfloat[caging (top view)] {\includegraphics[width=0.4\columnwidth]{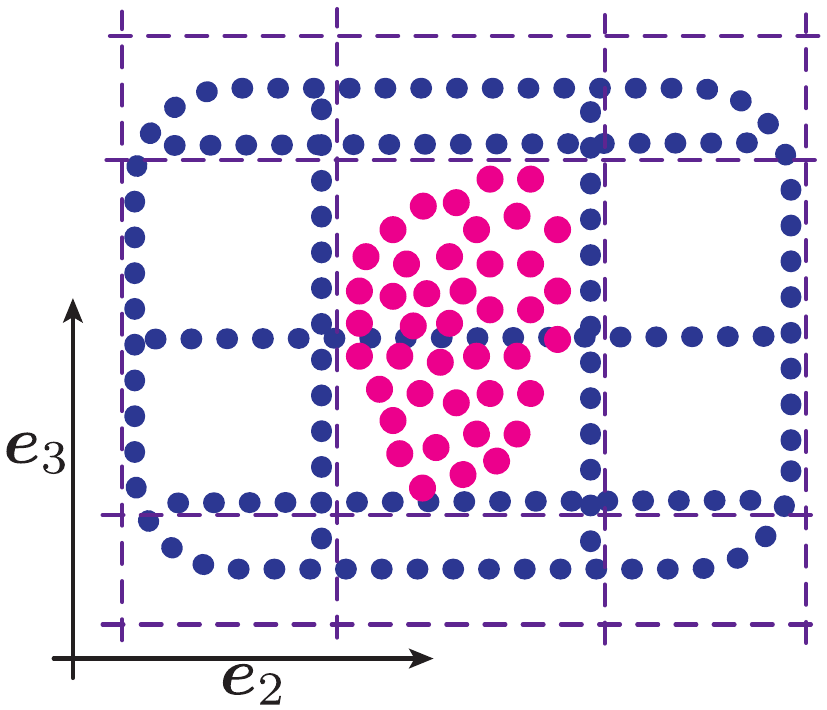}\label{fig:features_c}}
    \qquad
    \subfloat[signatures of geometry] {\includegraphics[width=0.46\columnwidth]{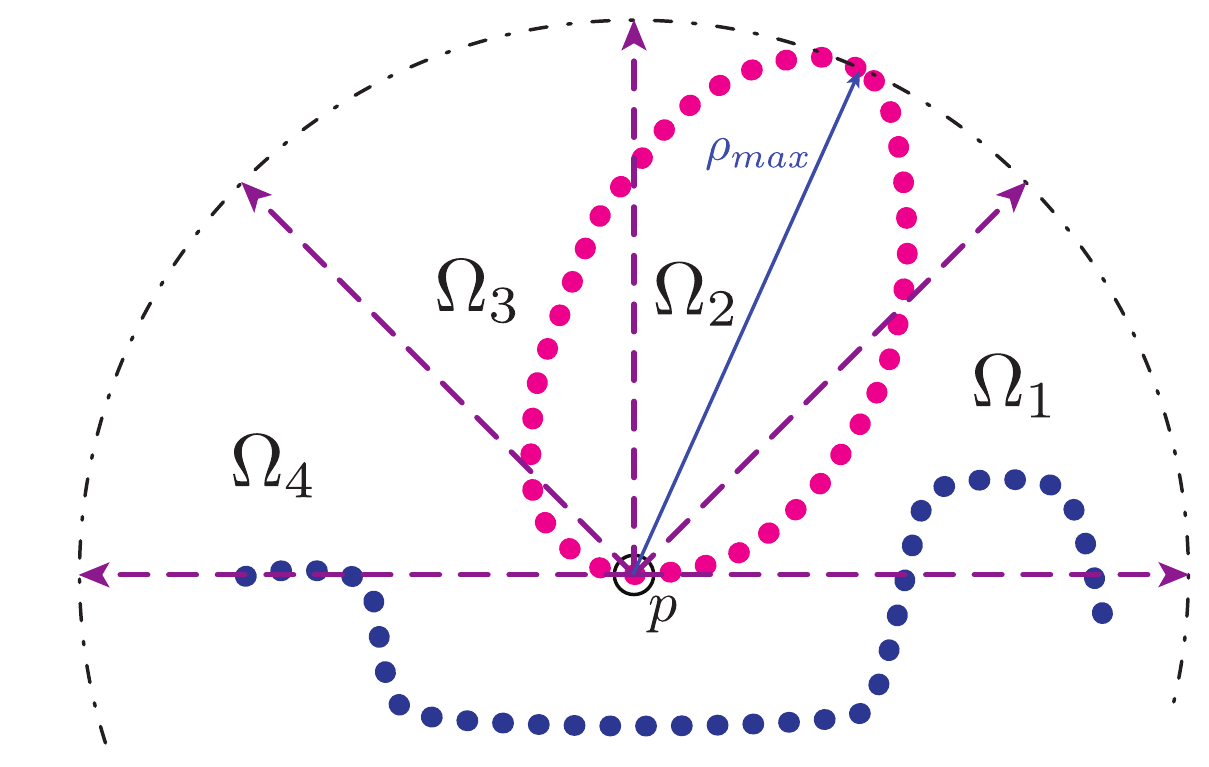}\label{fig:features_d}}
    \caption{{\small Illustration of Features in 2D.}}
    \label{fig:features}
\end{figure*}

%
\section{Learning Approach}\label{sec:learning}
\subsection{Features}\label{sec:feature}

In this section, we describe the features used in our learning algorithm. 
Given the point cloud of both object and placing area, we first randomly
sample some points in the bounding box of the environment as the placement candidates.
For each candidate, the features are computed to reflect the possibilities to place
the object there. In particular, we design the features to capture
the following two properties:

\begin{itemize}
\item \textbf{Supports and Stability.}
The object should be able to stay still after placing, and even better to be
able to stand small perturbations. 

\item \textbf{Preferred placements.}
Placements should also follow common practice. For example, the plates should be inserted into a dish-rack vertically
and glasses should be placed upside down on a stemware holder. 
%
\end{itemize}

In the following description, we use $\mathsf{O}$ to denote the set of object points,
$\bm{p}_o$ is the 3D coordinate of a point $o$ on the object, and $\bm{x}_t$ denotes
the coordinate of a point $t$ in the placing area point cloud.



\smallskip
\noindent\textbf{Supporting Contacts:}
We propose the features to reflect the support of the object from the placing 
environment. In particular, given an object represented by $n$ points and a possible placement, 
we first compute the vertical distance, $c_i$, $i$=$1$...$n$,
between each object point and placing area (Fig.~\ref{fig:features_a}), 
then the minimum $k$ distances are quantified into three features: 1) minimum distance $\min_{i=1\dots k} c_i$,
2) maximum distance $\max_{i=1\dots k}c_i$, and 3) the variance $\frac{1}{k}\sum_{i=1}^n (c_i - \bar{c})^2$, 
where $\bar{c} = \frac{1}{k}\sum_{i=1}^k c_i$.



\smallskip
\noindent\textbf{Caging:}
When the object is placed stably, not only is it supported by vertical contacts but
also it may lean against other local part of the environment and be ``caged'' by the
gravity and the surrounding environment.
Caging also ensures robustness to perturbations.
For instance, consider a pen placed upright in a holder. While it has only a
few vertical supporting contacts and may move because of a perturbation,
it will still remain in the holder because of caging.
To capture local caging shape, we divide the space
around the object into $3\times3\times3$ zones. The whole divided space is the
axis-aligned bounding box of the object scaled by $1.6$, and the center zone is
$1.05$ times of the bounding box (Fig.~\ref{fig:features_b} and \ref{fig:features_c}). The point cloud of
the placing area is partitioned into these zones labelled by 
$\Psi_{ijk}, i,j,k=1,2,3$,
where $i$ indexes the vertical direction $\bm{e}_1$, and $j$ and $k$ index the
other two orthogonal directions, $\bm{e}_2$ and $\bm{e}_3$, on horizontal plane.

From the top view, there are 9 regions (Fig.~\ref{fig:features_b}), each of which
covers three zones in the vertical direction.  For each region, the height of
the highest point in vertical direction is computed. This leads to 9 features.
In addition, we use the horizontal distance between environment and object to capture
the possible side support. In particular, for each $i=1,2,3$, we compute
\begin{equation}
\begin{split}
d_{i1} &= \min_{ \substack { \bm{x}_t\in\Psi_{i11}\cup\Psi_{i12}\cup\Psi_{i13} \\ \bm{p}_c\in\mathsf{O}} } {  \bm{e}_2^T(\bm{p}_o - \bm{x}_t)} \\
d_{i2} &= \min_{ \substack { \bm{x}_t\in\Psi_{i31}\cup\Psi_{i32}\cup\Psi_{i33} \\ \bm{p}_c\in\mathsf{O}} } { -\bm{e}_2^T(\bm{p}_o - \bm{x}_t)} \\
d_{i3} &= \min_{ \substack { \bm{x}_t\in\Psi_{i11}\cup\Psi_{i21}\cup\Psi_{i31} \\ \bm{p}_c\in\mathsf{O}} } {  \bm{e}_3^T(\bm{p}_o - \bm{x}_t)} \\
d_{i4} &= \min_{ \substack { \bm{x}_t\in\Psi_{i13}\cup\Psi_{i23}\cup\Psi_{i33} \\ \bm{p}_c\in\mathsf{O}} } { -\bm{e}_3^T(\bm{p}_o - \bm{x}_t)}
\end{split}
\end{equation}
and produce 12 additional features.


\smallskip
\noindent\textbf{Signatures of Geometry:}
A placement strategy in general depends on the geometric shapes of both the
object and the environment. To abstract the geometries, we propose the
signatures of point-cloud objects and environments and use them as
features in our learning algorithm. 

To compute the signatures of the object, we first compute the spherical
coordinates of all points with the origin at the placing point $\bm{p}$ (See
Fig.~\ref{fig:features_d}). Let $(\rho_i, \theta_i, \phi_i)$ denote the spherical
coordinate of point $i$. We then partition these points by their inclination
and azimuth angles. Given a spherical region $\Omega_{ab}, a=0\ldots3,b=0\ldots7$, a point $i$ is 
in $\Omega_{ab}$ when it satisfies $45a\le \theta_i \le 45(a+1)$ and $45b\le \phi_i\le 45(b+1)$.
This partition leads to 32 regions. We count the number of points in each region as a feature, 
creating 32 more features in total.

For the signatures of the environment, we conduct similar process, but consider only the local points
of the environment around the placing point $\bm{p}$. Let 
$$\rho_{max} = \max_{\textrm{object point $i$}}\rho_i.
$$
Only the environment points whose distance to $\bm{p}$ is less than $1.5\rho_{max}$ is partitioned
into the aforementioned 32 regions. This produces 32 more features.

To capture the matching between environment and object geometries, we first compute two 
values for each of the 32 regions $\Omega_{ab}$:
\begin{equation}
    \begin{split}
        t_{ab} &= \min_{\substack{ \textrm{enviroment point $i\in\Omega_{ab}$} \\ \rho_i\le1.5\rho_{max}}}\rho_i \\
        c_{ab} &= \max_{\textrm{object point $i\in\Omega_{ab}$}}\rho_i
    \end{split}
\end{equation}
and compute $c_{ab}/t_{ab}$ as a feature. Note that if there is no object or environment 
point in some region, we simply set a fixed number ($-1$ in practice) for the 
corresponding feature.

In total, we generate $120$ features: $3$ features for supporting contacts, 
$21$ features for caging, and $96$ of them for the signatures of geometry.
They are concatenated into a vector $\bm{v}\in\Re^{120}$, and used in the
learning algorithm described in the following sections.

%% file: algorithm.tex
\subsection{Learning Algorithm}\label{sec:algorithm}

We frame the manipulation task of placing as a supervised learning problem.
Given the features computed from the point-clouds of the object and the environment,
the goal of the learning algorithm is to
find a good placing hypothesis.

If we look at the objects and their placements in the environment, we notice
that there is an intrinsic difference between different placing settings.  
For example, it seems unrealistic to assume placing dishes into a rack and
hanging martini glasses upside down share exactly the same hypothesis, although
they might agree on a subset of attributes.   I.e., while some 
attributes may be shared across different objects and placing settings,
there are some attributes that are specific to the particular 
setting.
In such a scenario, it is not sufficient to have either one single model
or several completely independent models for each placing setting. The latter
also tends to suffer from over-fitting easily. Therefore, in this work we
propose to use shared sparsity structure in our learning model.

Say, we have $M$ objects and $N$ placing areas, thus making a total
of $r=MN$ placing settings.  We call each of these settings a `task'. 
Each task can have its own model but intuitively they should share parameters
underneath. To quantify this constraint, we use the idea from a recent work of
Jalali et al.~\cite{jalalidirty} that proposes to use a shared sparsity
structure for multiple linear regression. We extend their model to classic
soft-margin SVM~\cite{SVM}. 
(For specific details on experimental details and results, see
Section~\ref{sec:learning_scenario} and Section~\ref{sec:offline_experiments}.)

In detail, for $r$ tasks, let $X_i\in \Re^{p \times n_i}$ and $Y_i$ denote
training data and its corresponding label, where $p$ is the size of the feature set
and $n_i$ is the number of data points in task $i$. If we treat $r$ tasks
independently, we would get the following goal function based on classic SVM,
\begin{eqnarray}
\min_{\omega_i, b_i, i=1,\ldots, r} & \sum_{i=1}^r \left( \frac{1}{2} \left\|\omega_i\right\|^2_2 + C \sum_{j=1}^{n_i} \xi_{i,j} \right) \cr
\text{subject to} & Y_i^j(\omega_i^TX_i^j+b_i) \geq 1-\xi_{i,j}, \quad \xi_{i,j}\geq 0\cr
& \forall 1\leq i\leq r, \quad 1\leq j \leq n_i
\label{eq:svm}
\end{eqnarray}
where $\omega_i \in \Re^p$ is the learned model for $i$th task. $C$ is the trade off between the margin and training error, and $\xi_{i,j}$ is the slack variable.

Now, 
we modify the objective function above.  We
model each $\omega_i$ as composed of two parts $\omega_i = S_i + B_i$: 
the term $S_i$ represents the self-owned features and $B_i$ represents the shared features. 
All $S_i$'s should only have a few non-zero values so that it can reflect 
individual difference to some extent but would not become dominant in the final model. 
As for $B_i$, they need not have identical value, but should share similar sparsity
structure across tasks. I.e., for each feature, they should be either all very active or non-active
mostly. Let $\left\|S\right\|_{1,1} = \sum_{i,j}|S^j_i|$ and
$\left\|B\right\|_{1,\infty} = \sum_{j=1}^p \max_i |B_i^j|$. Our
new goal function is now:
\begin{eqnarray}
\min_{\omega_i, b_i, i=1,\ldots, r} &  \sum_{i=1}^r \left( \frac{1}{2} \left\|\omega_i\right\|^2_2 + C \sum_{j=1}^{n_i} \xi_{i,j} \right) + \cr
& \qquad \quad \lambda_S \left\|S\right\|_{1,1} + \lambda_B \left\|B\right\|_{1,\infty}\cr
\text{subject to} & Y_i^j(\omega_i^TX_i^j+b_i) \geq 1-\xi_{i,j}, \quad \xi_{i,j}\geq 0\cr
& \qquad \forall 1\leq i\leq r, 1\leq j \leq n_i
\label{eq:dirty}
\end{eqnarray}
This function contains two penalty terms for $S$ and $B$ each, with
hand-tuned coefficients $\lambda_S$ and $\lambda_B$. Because
$\left\|S\right\|_{1,1}$ is defined as the sum of absolute values of elements
in each $S_i$, it can effectively control the magnitude of $S$ without
interfering with the internal structure of $S$. For sharing,
$\left\|B\right\|_{1,\infty}$ encourages all $B_i$ to simultaneously assign large
weight (can be either positive or negative) to the same set of features. This is
because no additional penalty is added for increasing $B_i^j$ if it is not already the
maximum one.
This modification indeed results in a superior performance with new objects
in new placing areas. 


We transform this optimization problem into a standard quadratic programming
(QP) problem by introducing auxiliary variables to substitute for the absolute and the maximum value
terms.  Unlike Equation~\ref{eq:svm} which decomposes into $r$ sub-problems,
the optimization problem in Equation~\ref{eq:dirty} becomes larger, and hence
takes a lot of computational time to learn the parameters.  
However, inference is still fast since predicting the score is simply the dot product of the features
with the learned parameters. During the test, if the task is in the training set, then its corresponding model is
used to compute the scores. Otherwise we use a  voting system, in which we average of the scores
from all the models in the training data to predict score for
the new situation (see Section~\ref{sec:learning_scenario} for  different training settings). 






%% file: experiment.tex
\section{Experiments}\label{sec:experiments}

\subsection{Robot Hardware and Setup}

We use a Adept Viper s850 arm with six degrees of freedom, equipped with a
parallel plate gripper that can open to up to a maximum width of 5cm. The arm
has 
a reach of 105cm, together with our gripper. The
arm plus gripper has a repeatability of about 1mm in XYZ positioning, but there is no force or tactile feedback in our arm.
We use a Bumblebee\footnote{http://www.ptgrey.com/products/stereo.asp} camera to obtain
the point clouds.

\subsection{Learning Scenarios\label{sec:learning_scenario}}

In real-world placing, the robot may or may not have seen the placing
locations (and the objects) before. Therefore, we train our algorithm for four
different scenarios:
\begin{enumerate}
\item 
Same Environment Same Object (\textbf{SESO}),
\item 
Same Environment New Object (\textbf{SENO}),  
\item 
New Environment Same Object (\textbf{NESO}), 
\item 
New Environment New Object (\textbf{NENO}).  
\end{enumerate}

If the environment is `same', then only this environment is included in the training data, otherwise 
all other environments are included except the one for test. This is done similarly for the objects.
Through these four scenarios, we would be able to observe the algorithm's performance thoroughly. 

We also compare our algorithm with  the following three heuristic methods:

\begin{enumerate}
\item \textbf{Chance.} This method randomly samples a ``collision-free'' location (from the 
bounding box of the environment) and an orientation for placing.

\item
\textbf{Flat surface upright rule.}
Several methods exist for finding `flat' surfaces \cite{kemp-placing}, and we consider a 
placing method based on finding flat surfaces.
In this method, objects would be placed with pre-defined upright orientation on the surface when 
flat surfaces exist in a placing area such as a table or a pen holder.
When no flat surface can be found, such as for racks or stemware holder, 
this method would pick placements randomly.

\item
\textbf{Finding lowest placing point.} For many placing areas,
    such as dish-racks or containers, a lower placing point (see Section~\ref{sec:feature})
    often gives more stability. Therefore,
   this heuristic rule chooses the placing point with the lowest height.

\end{enumerate}

\subsection{Evaluation Metrics}

We evaluate our algorithm's performance on the following metrics:
\begin{itemize}
\item $R_0$: Rank of the first valid placement. 
($R_0=1$ in the best case)
\item \textit{Precision@n}: In top $n$ candidates, the fraction of valid placements. 
Specifically, we choose $n=5$ in our evaluations. ($0 \leq {\rm Pre@n} \leq 1$.)
\item $P_\text{stability}$: Success-rate (in \%) of robotic placement in placing the 
object stably, i.e., the object does not move much after placing.
\item $P_\text{preference}$: Success-rate (in \%) of robotic placements while counting
even stable placements as incorrect if they are not the `preferred' ones.
\end{itemize}
Except $R_0$, the other three metrics represent precision and thus higher values
 indicate higher performance.

\begin{table}[t!]
\centering
\caption{{\small Average performance of the algorithm when used with different
types of features.
Tested on SESO scenario with independent SVMs.}}
\label{tbl:features}
\begin{tabular}{l c c c c c}
\toprule%
	&	chance	&	contact	&	caging	&	signature	& all		\\
\midrule
$R_0$	&	29.4	&	13.3	&	5.0	&	2.6	&	\bf{1.0}	\\
Pre@5	&	0.10	&	0.64	&	0.69	&	0.82	&	\bf{0.96}	\\
\bottomrule%
\end{tabular}
\end{table}

\begin{table}[t!]
\centering
\caption{{\small Average performance for different training methods: joint SVM, independent SVM with voting and shared sparsity SVM with voting for NENO scenario.}}
\label{tbl:alg}
\begin{tabular}{l c c c c}
\toprule%
	&	chance	&	joint 	&	independent	&	shared	\\
\midrule    
$R_0$	&	29.4	&	9.4	&	5.3	&	\bf{1.9}	\\
Pre@5	&	0.10	&	0.54	&	0.61	&	\bf{0.66}	\\
\bottomrule%
\end{tabular}
\end{table}

\begin{table*}[htb!]
\caption{\small {\bf Learning experiment statistics:}  The performance of different learning algorithms in different scenarios
is shown. 
The first three double columns 
are the results for baselines, where no training data is used. Columns under `independent SVM' are trained using separate classic SVM on each task, under
four learning scenarios. The last double column is trained via shared sparsity SVM only for NENO. For the particular case of the martini glass and stemware holder, marked by `-',
statistics for SENO are not available because no other objects can be well placed in this environment. 
 \label{tbl:offline_results}}
{\footnotesize
\newcolumntype{P}[2]{>{\footnotesize#1\hspace{0pt}\arraybackslash}p{#2}}
\setlength{\tabcolsep}{2pt}
\centering
\resizebox{\hsize}{!} {
\begin{tabular}{p{0.1\linewidth}p{0.18\linewidth}|P{\centering}{7mm}P{\centering}{8mm}P{\centering}{7mm}P{\centering}{8mm}P{\centering}{7mm}P{\centering}{8mm}|P{\centering}{7mm}P{\centering}{8mm}P{\centering}{7mm}P{\centering}{8mm}P{\centering}{7mm}P{\centering}{8mm}P{\centering}{7mm}P{\centering}{8mm}|P{\centering}{7mm}P{\centering}{8mm}}
\multicolumn{18}{c}{Listed environment-wise, averaged over the objects.} \\
\whline{1.1pt} 
& & \multicolumn{6}{c|}{heuristic placement} & \multicolumn{8}{c|}{independent SVM with voting} &\multicolumn{2}{c}{ {\relsize{-1}shared sparsity} }\\
\cline{3-18}
& & \multicolumn{2}{c}{chance} & \multicolumn{2}{c}{flat surface} & \multicolumn{2}{c|}{lowest pt} & \multicolumn{2}{c}{SESO} &  \multicolumn{2}{c}{SENO} & \multicolumn{2}{c}{NESO} & \multicolumn{2}{c|}{NENO} & \multicolumn{2}{c}{NENO} \\
\whline{0.4pt} 
     environment & objects & $R_0$ & Pre@5 & $R_0$ & Pre@5 & $R_0$ & \multicolumn{1}{c|}{Pre@5} & $R_0$ & Pre@5 & $R_0$ & Pre@5 & $R_0$ & Pre@5 & $R_0$ & \multicolumn{1}{c|}{Pre@5} & $R_0$ & Pre@5 \\ 
\whline{0.8pt} 
rack-1 	&    plate,mug,martini,bowl                                      	& 3.8	& 0.15	& 1.8	& 0.50	& 1.3	& 0.75	& 1.0	& 0.85	& 1.0	& 0.90	& 1.3	& 0.85	& 2.3	& 0.75	& 1.3	& 0.70 \\
rack-2 	&    plate,mug,martini,bowl                                      	& 5.0	& 0.25	& 18.3	& 0.05	& 22.0	& 0.10	& 1.0	& 0.90	& 1.5	& 0.75	& 2.5	& 0.55	& 4.8	& 0.60	& 1.5	& 0.55 \\
rack-3 	&    plate,mug,martini,bowl                                      	& 4.8	& 0.15	& 4.8	& 0.20	& 22.0	& 0.15	& 1.0	& 1.00	& 3.8	& 0.40	& 5.8	& 0.35	& 6.3	& 0.50	& 3.8	& 0.40 \\
flat   	&    plate,mug,martini,bowl, candy cane, disc, spoon, tuning fork 	& 6.6	& 0.08	& 1.0	& 0.98	& 4.3	& 0.23	& 1.0	& 1.00	& 1.1	& 0.98	& 2.8	& 0.85	& 1.1	& 0.98	& 1.4	& 0.88 \\
pen holder  	&    candy cane, disc, spoon, tuning fork                    	& 128.0	& 0.00	& 61.3	& 0.05	& 60.3	& 0.60	& 1.0	& 1.00	& 1.8	& 0.75	& 3.8	& 0.40	& 3.8	& 0.30	& 1.3	& 0.80 \\
hook   	&    candy cane, disc                                            	& 78.0	& 0.00	& 42.0	& 0.00	& 136.5	& 0.00	& 1.0	& 1.00	& 2.5	& 0.40	& 5.0	& 0.10	& 8.5	& 0.05	& 2.5	& 0.60 \\
stemware holder    	&    martini                                         	& 18.0	& 0.00	& 65.0	& 0.00	& 157.0	& 0.00	& 1.0	& 1.00	& -	& -	& 45.0	& 0.00	& 50.0	& 0.00	& 4.0	& 0.20 \\
\whline{0.6pt} 
\multicolumn{2}{c|}{\bf{Average}} 		& \bf{29.4}		& \bf{0.10}		& \bf{18.6}		& \bf{0.41}		& \bf{32.8}		& \bf{0.30}		& \bf{1.0}		& \bf{0.96}		& \bf{1.8}		& \bf{0.76}		& \bf{4.8}		& \bf{0.58}		& \bf{5.3}		& \bf{0.61}		& \bf{1.9}		& \bf{0.66}\\
\whline{0.8pt} 
\\
\multicolumn{18}{c}{Listed object-wise, averaged over the environments.}\\
\whline{0.8pt} 
object & environments & $R_0$ & Pre@5 & $R_0$ & Pre@5 & $R_0$ & \multicolumn{1}{c|}{Pre@5} & $R_0$ & Pre@5 & $R_0$ & Pre@5 & $R_0$ & Pre@5 & $R_0$ & \multicolumn{1}{c|}{Pre@5} & $R_0$ & Pre@5 \\ 
\whline{0.8pt} 
plate   	&    rack-1, rack-2, rack-3, flat               	& 4.0	& 0.20	& 4.3	& 0.45	& 27.5	& 0.25	& 1.0	& 0.95	& 2.3	& 0.60	& 4.8	& 0.55	& 5.0	& 0.55	& 3.0	& 0.55 \\
mug     	&    rack-1, rack-2, rack-3, flat               	& 5.3	& 0.10	& 11.8	& 0.50	& 3.8	& 0.35	& 1.0	& 0.95	& 2.0	& 0.75	& 3.8	& 0.50	& 1.0	& 0.90	& 1.8	& 0.55 \\
martini	&  rack-1, rack-2, rack-3, flat, stemware holder  	& 6.8	& 0.12	& 16.0	& 0.32	& 39.0	& 0.32	& 1.0	& 1.00	& 1.0	& 1.00	& 12.4	& 0.56	& 10.8	& 0.76	& 2.0	& 0.56 \\
bowl    	&    rack-1, rack-2, rack-3, flat               	& 6.5	& 0.15	& 6.0	& 0.40	& 7.0	& 0.30	& 1.0	& 0.85	& 2.0	& 0.70	& 1.0	& 0.75	& 7.3	& 0.45	& 1.8	& 0.70 \\
candy cane  	&    flat, pen holder, hook                  	& 102.7	& 0.00	& 44.0	& 0.33	& 51.7	& 0.33	& 1.0	& 1.00	& 1.7	& 0.73	& 3.7	& 0.40	& 3.3	& 0.43	& 1.7	& 0.73 \\
disc    	&    flat, pen holder, hook                      	& 32.7	& 0.00	& 20.0	& 0.40	& 122.7	& 0.00	& 1.0	& 1.00	& 2.0	& 0.67	& 2.3	& 0.67	& 5.7	& 0.40	& 1.3	& 0.87 \\
spoon   	&    flat, pen holder                            	& 101.0	& 0.20	& 35.0	& 0.50	& 2.5	& 0.50	& 1.0	& 1.00	& 1.5	& 0.90	& 4.5	& 0.40	& 3.5	& 0.50	& 1.5	& 0.90 \\
tuning fork  	&    flat, pen holder                        	& 44.0	& 0.00	& 35.5	& 0.40	& 5.0	& 0.50	& 1.0	& 1.00	& 1.5	& 0.80	& 1.5	& 0.80	& 1.5	& 0.80	& 1.5	& 0.80 \\
\whline{0.6pt} 
\multicolumn{2}{c|}{\bf{Average}} 	& \bf{29.4}	& \bf{0.10}	& \bf{18.6}	& \bf{0.41}	& \bf{32.8}	& \bf{0.30}	& \bf{1.0}	& \bf{0.96}	& \bf{1.8}	& \bf{0.76}	& \bf{4.8}	& \bf{0.58}	& \bf{5.3}	& \bf{0.61}	& \bf{1.9}	& \bf{0.66}\\
\whline{1.1pt} 
\end{tabular}
}
}
\end{table*}

\subsection{Learning Experiments\label{sec:offline_experiments}}
For evaluating our learning algorithm, we considered 7 environments and 8 objects
(shown in Fig.~\ref{fig:4_scenes}).
In detail, we generated one dataset each for
training and test for each setting (i.e., an object-environment pair). In each dataset, 
100 distinct 3D locations are paired with 18 different orientations which gives 1800 different placements.
After eliminating placements that have collisions, we have
37655 placements in total.

Table~\ref{tbl:features} shows the average performance when we use
different types of features: supporting contacts, caging
and geometric signatures. While all the three types of features outperform
chance, combining them together gives the highest $R_0$ and Pre@5. 
Next, Table~\ref{tbl:alg} shows the comparison of three variations of SVM learning 
algorithms: 1) joint SVM where one single model is learned from all the placing settings
in the training dataset;
2) independent SVM that treats each setting as a learning task and learns separate model 
for every setting;
3) shared sparsity SVM that also learns one model per setting but with parameter sharing.
Both independent and shared sparsity SVM  use voting to rank placements for the test case.
Table~\ref{tbl:alg} shows
that in the hardest learning scenario, NENO, the shared sparsity SVM performs
best. The result also indicates that independent SVM with voting is better than
joint SVM. This could be due to the large variety in the placing situations
in the training set.
 Thus imposing one model for all tasks  
 decreases the performance.

Table~\ref{tbl:offline_results}
shows the performance of the different algorithms described in 
Section~\ref{sec:learning_scenario} on various placing tasks.
For each row in Table~\ref{tbl:offline_results}, the numbers are averaged across the objects for
each environment (when listed environment-wise) and are averaged across the environments for
each object (when listed object-wise).

There is a large variety in the objects as well as in 
the environments, leading to a large number of possible placements. 
Thus one can hardly find a heuristic that would find valid placements in all the 
cases.  Not surprisingly, the chance method performs poorly (Prec@5=0) because 
there are very few  preferred placements in the large sampling space of possible placements.
The two heuristic methods perform well in some obvious cases, e.g., flat-surface-upright method works
well for flat surfaces, and lowest-point method works reasonably in `cage-like' environments  
such as a pen holder. However,  their performance varies significantly in non-trivial cases.
They perform poorly in many cases including the stemware holder and the hook.

We get close to perfect results for SESO case---i.e., the learning algorithm
can very reliably predict object placements if a known object was being 
placed in a previously seen location.   The performance is still very high
for SENO (i.e., even if the robot has not seen the object before), where the first correct 
prediction is ranked 1.8 on average. This means we only need to perform simulation
twice.  However, for NESO the performance starts to deteriorate---the
average number of simulations needed is 4.8 because of poor performance in 
placing the martini glass on the flat surface and the stemware holder. 

The last learning scenario, NENO is extremely challenging---here, for each
object/placing area pair, the algorithm is trained without either the object
or the placing area. With the same algorithm, independent SVM with voting, 
$R_0$ increases to 5.3. However, shared sparsity SVM 
(the last column in the table) helps to reduce the average $R_0$ to 1.9.
It is worth noting that in cases where the placing strategy is very different from the ones trained on, 
our algorithm does not perform well, e.g., $R_0$ is 4.0 for the martini glass and stemware holder.
This issue can be potentially addressed by expanding the training dataset.

\begin{table}[t!]

\centering
\caption{{\small Robotic performance test results, trained using shared sparsity SVM under different learning scenarios: SESO and
NENO. Five experiments each were performed for each object-placing area pair. $P_s$ stands for $P_\text{stability}$ and $P_p$ stands for $P_\text{preference}$.}
} \label{tbl:online_results}

{\footnotesize
\begin{tabular}{l l ccc ccc}
    \toprule
        & & \multicolumn{3}{c}{SESO} &  \multicolumn{3}{c}{NENO}\\
    \cmidrule{3-8}
		environment & object & $R_0$ & $P_s$ & $P_p$ & $R_0$ & $P_s$ & $P_p$ \\ 
        & & & (\%)& (\%)& & (\%)& (\%)\\
	\toprule
\multirow{3}{*}{rack-1}	&	plate	&	1.0	&	100	&	100	&	5.8	&	100	&	60\\
	&	martini	&	1.0	&	100	&	100	&	1.0	&	100	&	100\\
	&	bowl	&	1.0	&	100	&	100	&	9.4	&	100	&	80\\
\hline
\multirow{3}{*}{rack-3}	&	plate	&	1.0	&	100	&	100	&	1.0	&	100	&	100\\
	&	martini	&	1.0	&	100	&	100	&	1.0	&	100	&	100\\
	&	bowl	&	1.0	&	80	&	80	&	1.0	&	100	&	100\\
\hline
\multirow{3}{*}{flat}	&	plate	&	1.0	&	100	&	100	&	2.0	&	100	&	100\\
	&	martini	&	1.0	&	100	&	100	&	1.0	&	100	&	100\\
	&	bowl	&	1.0	&	100	&	100	&	1.8	&	100	&	100\\
\hline
stemware &	martini	&	1.0	&	100	&	100	&	1.0	&	80	&	80\\
holder & & & & & & &\\             
\midrule
\multicolumn{2}{c}{\bf Average}	&\bf 1.0	&	\bf{98}	&	\bf{98}	&	\bf{2.5}	&	\bf{98}	&	\bf{92}\\
	\bottomrule
	\end{tabular}
}    
\end{table}

\begin{figure*}[t!]
\centering
\subfloat{
\includegraphics[height=0.15\textheight]{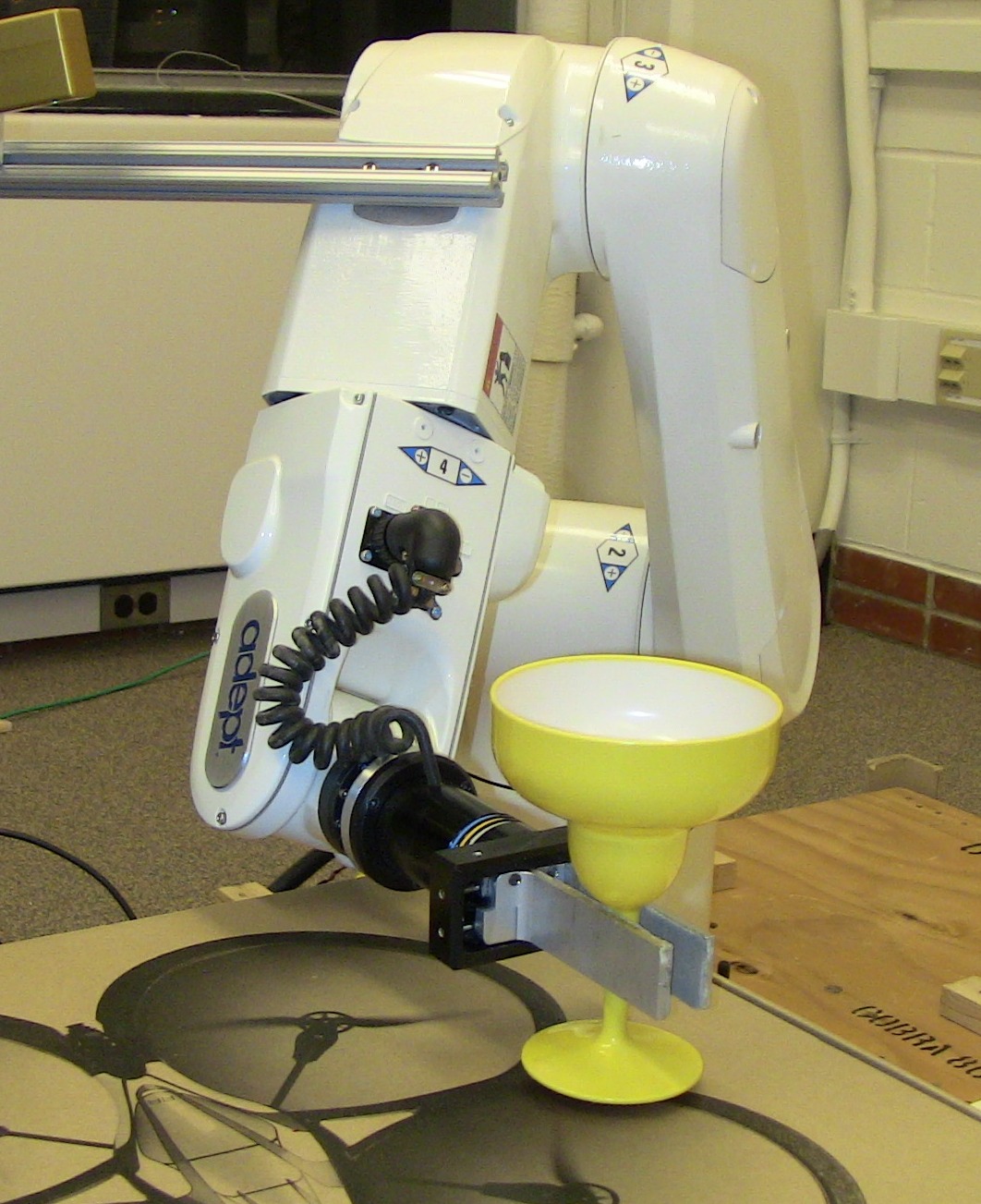}
\includegraphics[angle=90, height=0.15\textheight]{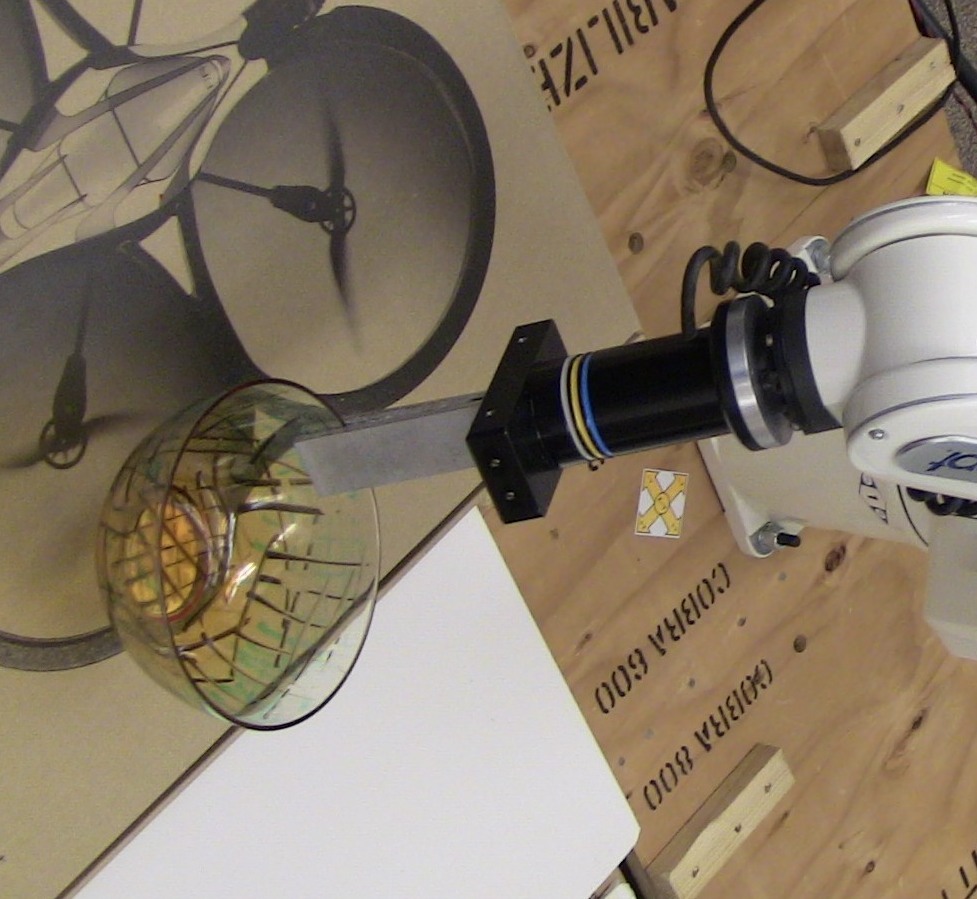}
\includegraphics[height=0.15\textheight]{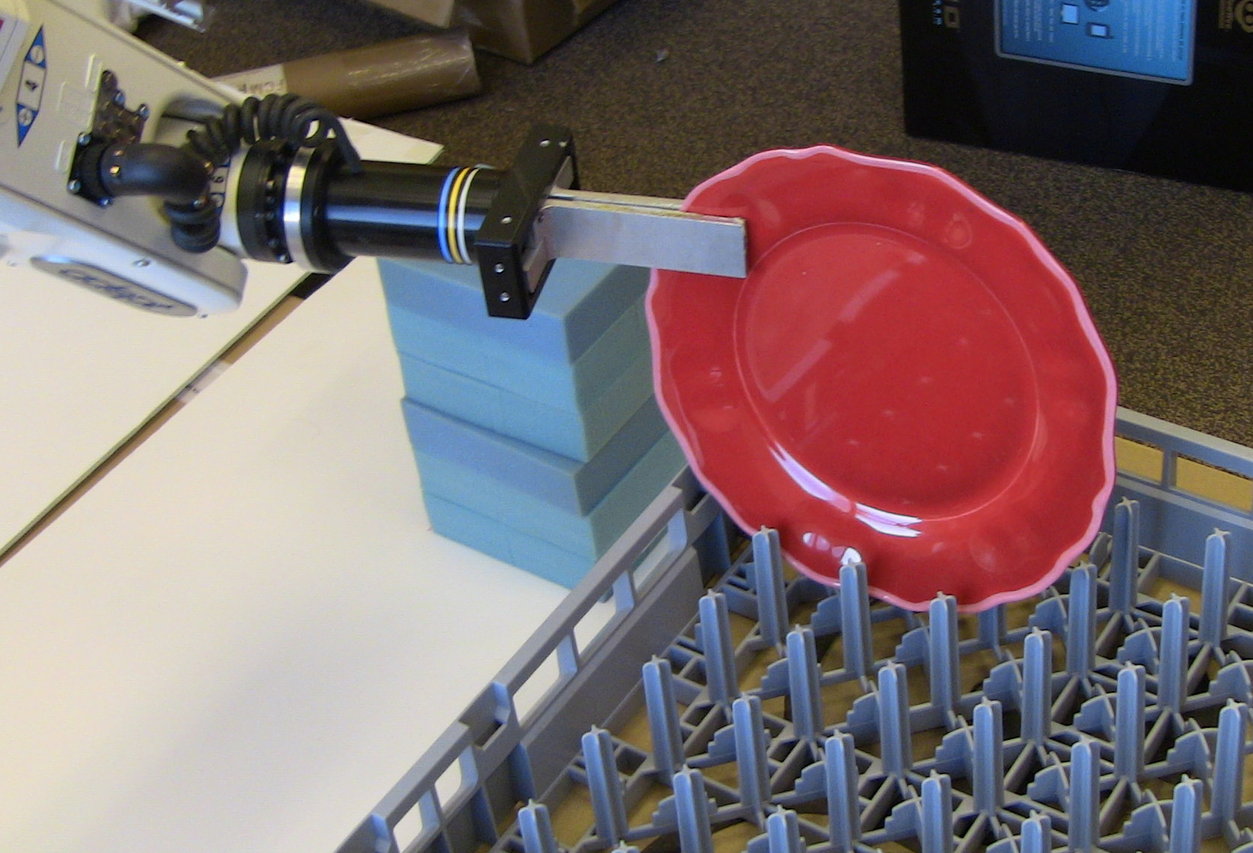}
\includegraphics[height=0.15\textheight]{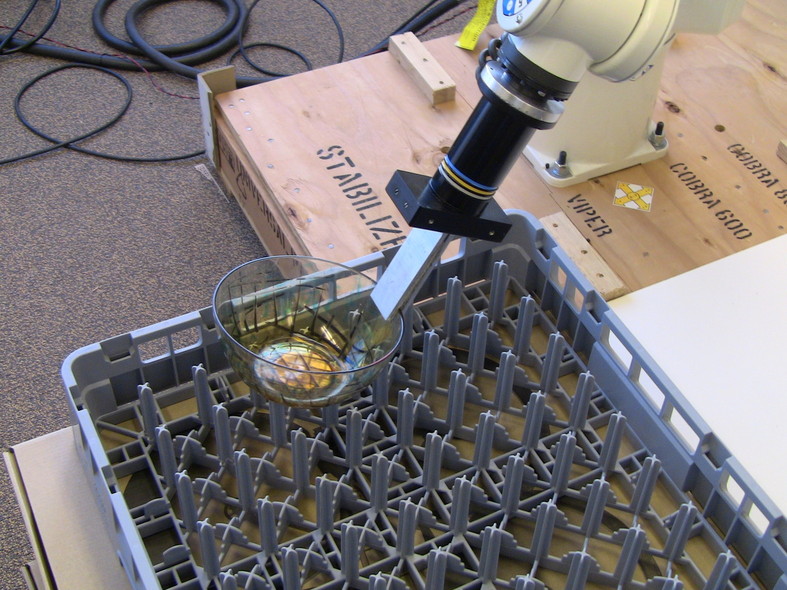}}\\
\hskip -0.02in
\subfloat{
\includegraphics[height=0.145\textheight]{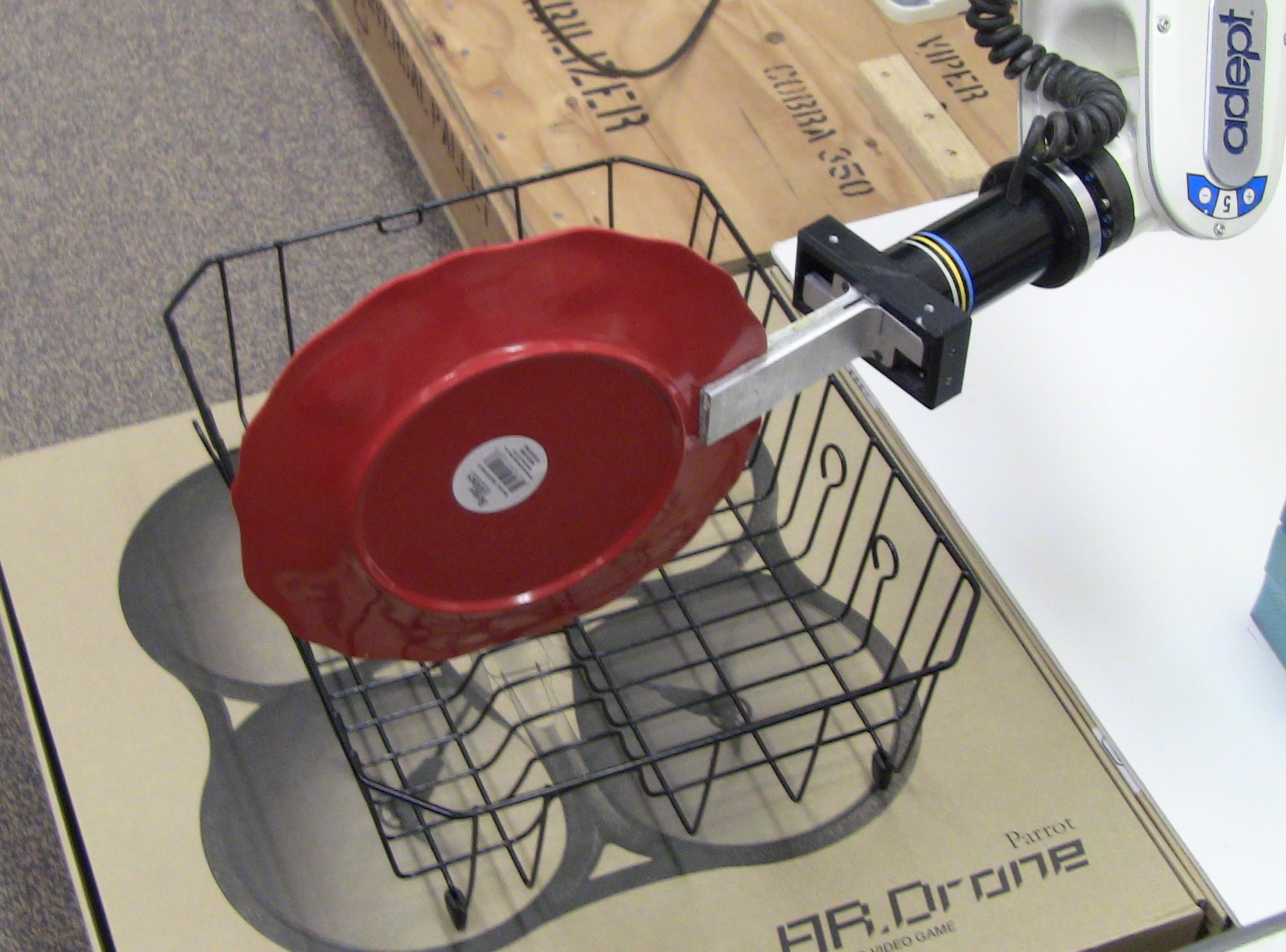}
\includegraphics[height=0.145\textheight]{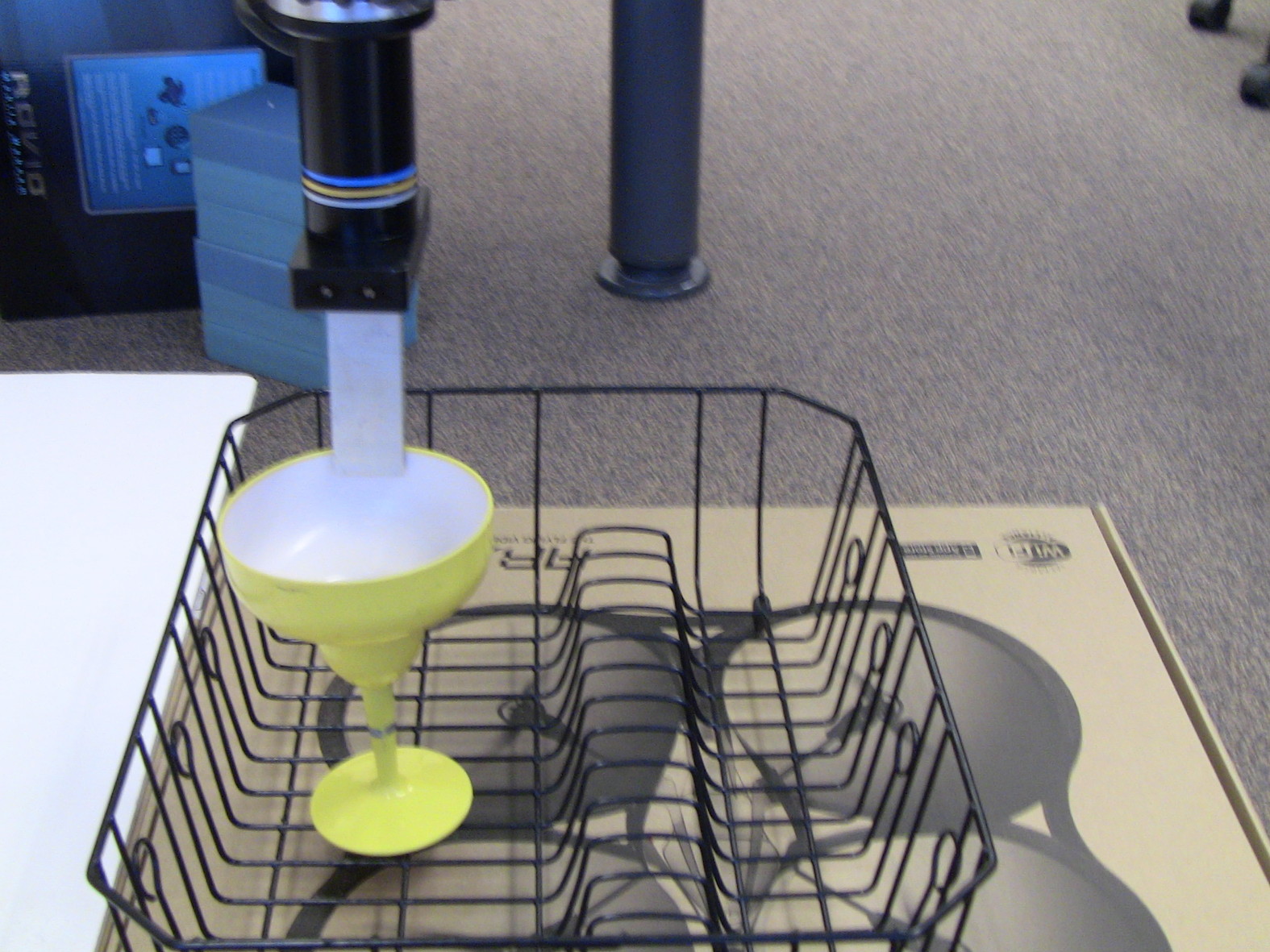}
\includegraphics[height=0.145\textheight]{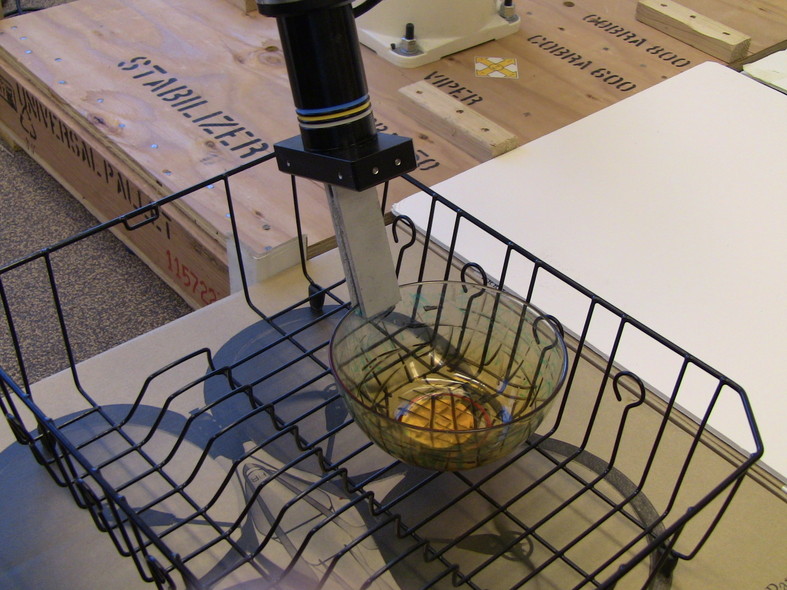}
\includegraphics[angle=90, height=0.145\textheight]{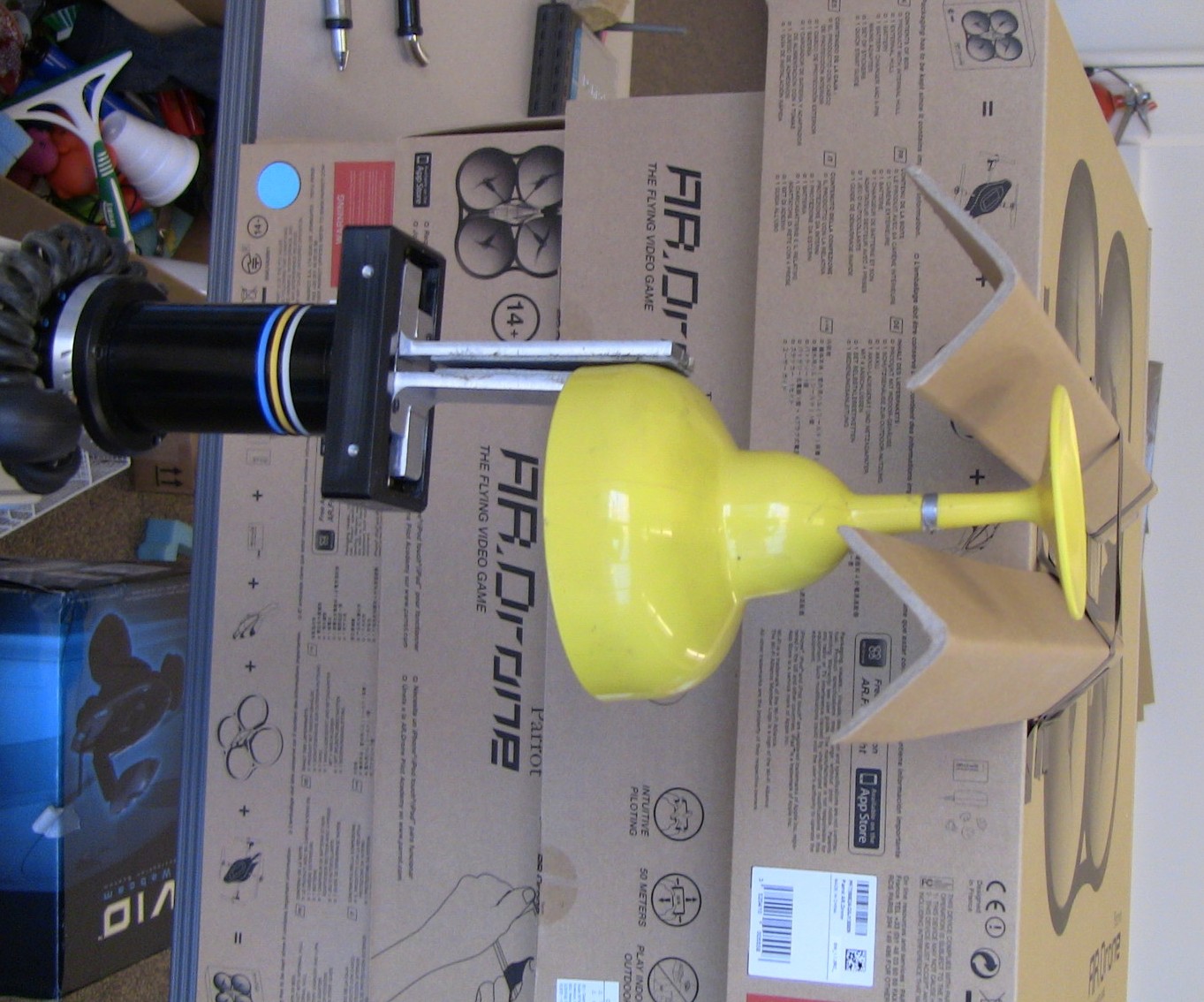}}
\caption{{\small Some screenshots of our robot placing different objects in several placing areas.
In the examples above, the robot placed the objects in stable as well as preferred
orientation, except for the top right image where
a bowl is placed stably in upright orientation on rack-1. However, a more preferred 
orientation is upside down.
}}
\label{fig:screenshots}
\end{figure*}

Note that for placing the objects in a designated placing area,
our method relies on learning algorithms trained from data instead
of relying on hard-coded rules.  The assumption about the pre-defined 
object-specific grasping locations can be eliminated by 
other grasping algorithms, e.g., \cite{jiangICRA}. We believe that this 
would enable our approach to extend to different and 
new placing scenarios.

\subsection{Robotic Experiments}

We conducted experiments on our robotic arm with
the system as described in Section~\ref{sec:system}. For training,
we used the same dataset as for the learning experiments in 
previous section. We performed a total of $100$ trials.

Table~\ref{tbl:online_results} shows results for the objects being placed by the 
robotic arm in four placing scenarios: flat surface, rack-1, rack-3 and stemware holder. 
We see that our SESO case obtains a 98\% success rate in placing the objects,
which is quite significant and shows the performance of our overall system.
It failed only in one experiment, when the bowl slid from the bump of the
rack because of a small displacement. 


In our NENO case, we obtain 98\% performance if we consider only stable
placing, but 92\% performance if we disqualify those stable placements that are
not the preferred ones. The plate is such an object for which these two
success-rates are quite different---since the learning algorithm has never seen
the plate before (and in fact the placing area either), it often predicts a
slanted or horizontal placement, which even though stable is not the preferred
way to place a plate in a dish-rack. One failure case was caused by
an error that occurred during grasping---the martini glass slipped a bit in the
gripper and thus could not fit into the narrow stemware holder.

Fig.~\ref{fig:screenshots} shows several screenshots of our robot
placing objects.
Some of the cases are quite tricky, for example placing the martini-glass
hanging from the stemware holder.  Videos of our robot placing
these objects is available at:
\texttt{http://pr.cs.cornell.edu/placingobjects}

%% file: conclusion.tex
\section{Conclusion and Discussion}
\label{sec:conclusion}
In this paper, we considered the problem of placing objects in various
types of placing areas, especially the scenarios when the objects and the 
placing areas may not have been seen by the robot before. We first
presented features that contain information about stability
and preferred placements. We then used a learning approach based on SVM with shared sparsity structure
for predicting the top placements.   
In our extensive learning and robotic experiments,
we show that different objects can be placed successfully in several
environments, including flat surfaces, several dish-racks, and hanging objects
upside down on a stemware holder and a hook.

There still remain several issues for placing objects that we have not addressed yet,
such as considering the reachability of the placements or performing more complicated 
manipulations during placing. 
Furthermore, we also need to consider detecting the appropriate placing areas for an object,
placing in cluttered scenes,  and how to place multiple objects 
in order for a personal robot to complete the tasks of interest such as arranging a disorganized kitchen or
cleaning a cluttered room.
